# Improving Electrolyte Performance for Target Cathode Loading Using Interpretable Data-Driven Approach

Vidushi Sharma*, Andy Tek[+], Khanh Nguyen[+], Max Giammona, Murtaza Zohair, Linda Sundberg and Young-Hye La

IBM Almaden Research Center, San Jose, CA 95120

*Email: vidushis@ibm.com

[+]Equal contribution by authors

## Abstract

Higher loading of active electrode materials is desired in batteries, especially those based on conversion reactions, for enhanced energy density and cost efficiency. However, increasing active material loading in electrodes can cause significant performance depreciation due to internal resistance, shuttling, and parasitic side reactions, which can be alleviated to a certain extent by a compatible design of electrolytes. In this work, a data-driven approach is leveraged to find a high-performing electrolyte formulation for a novel interhalogen battery custom to the target cathode loading. An electrolyte design consisting of 4 solvents and 4 salts is experimentally devised for a novel Li-ICl battery based on a multi-electron redox reaction. The experimental dataset with variable electrolyte compositions and active cathode (LiI) loading, is used to train a graph-based deep learning model mapping changing variables in the battery's material design to its specific capacity. The trained model is used to further optimize the electrolyte formulation compositions for enhancing the battery capacity at a target cathode loading by a two-fold approach: large-scale screening and interpreting electrolyte design principles for different cathode loadings. The data-driven approach is demonstrated to bring about an additional 20% increment in the specific capacity of the battery over capacities obtained from the experimental optimization. The approach resulted in an electrolyte with a high specific capacity of 250 mAh/g (at 1 mA/cm$^2$) and excellent rate capability at the targeted 45 wt% cathode loading, demonstrating the scope of a data-driven approach in improving battery electrolytes based on additional cell-level variables.

## 1. Introduction

Data-driven methods have significantly accelerated the material discovery cycle by enhancing the efficiency of various predictive [1, 2], generative [3, 4], and optimization processes [5, 6]. Machine learning (ML) models can process vast amounts of data and extract insights that can guide the development of novel materials for various fields such as pharmaceuticals [7], energy storage [8, 9], and electronics [2, 10]. Especially in the field of energy storage where the rapidly growing demands of the electric vehicle (EV) market have pushed the electrochemical energy storage industry to seek new battery material with high-performance capabilities, data-driven models like ML and generative artificial intelligence (AI) are proving to be invaluable tools in screening stable electrode designs cost-effectively and suggesting new materials with desired attributes [8, 9, 11, 12]. Despite these significant strides in the computational discovery of new battery materials, cases of their successful demonstration are scarce due to complex interactions between different battery components. This necessitates further development of these ML models to capture multi-constituent materials efficiently, enabling competent design optimization and performance prediction.

Conversion batteries combining iodine-based ($I_2$) cathodes, a non-aqueous electrolyte, and lithium (Li) metal anode are of significant interest due to the demonstrated high-rate capability of the cathode with the potential for ultra-fast charging speeds (>10C), good theoretical capacity and voltage profile (211 mAh/g, 3V) [13]. However, attempts to produce a commercially viable battery based on the iodine chemistry have been hindered thus far due to several depreciating phenomena. These include electron shuttling during charging, self-discharge, and parasitic reactions between the solubilized catholyte and the Li metal anode [14, 15]. The solubility of the active cathode material in the electrolyte has also limited the relative weight percentage (wt %) of active material in the cathode to lower than 50 wt % [16, 17], thereby, limiting the energy density of full cells [13]. These issues can be likely addressed to an extent with an appropriate electrolyte design for these emerging battery systems [18-20]. Except, formulating efficient electrolytes has been a significant challenge due to being a complex multi-constituent design space. Electrolytes typically constitute one major organic solvent mixed with an inorganic salt, together facilitating $Li^+$ ion transport and the formation of a passivating solid electrolyte interface (SEI) on the anode surface. New co-solvents and additives are frequently added to this mixture to expedite charging and stabilize interfaces against side reactions. For instance, the lithium nitrate ($LiNO_3$) additive is widely included in electrolytes as it promotes the spontaneous formation of a stabilizing SEI on the surface of the Li metal anode, thereby enhancing cycle stability [15]. Similarly, recent studies have shown that the incorporation of additional lithium chloride (LiCl) salt in iodine battery electrolytes can double the theoretical capacity from 211 mAh/g to 422 mAh/g by introducing additional multi-electron redox reaction leading to the formation of interhalogen species I-Cl [21]. These performance-enhancing additions to the electrolyte intensify their complexity and bring forth the necessity to optimize this mixture for compatibility with other battery components.

Electrolyte constituents and their relative compositions create a high-dimensional design space that makes finding optimum electrolyte formulation a 'needle-in-a-haystack' problem. Until recently, the issue of identifying suitable electrolyte constituents and composition optimization have been tackled separately, where computational simulations drive the screening of constituent salts and solvents based on their physiochemical properties [22-24], and stochastic machine learning (ML) methods such as Bayesian optimization (BO) drive composition optimization of a constrained chemical space, based on sizable electrolyte property datasets that are difficult to procure [25, 26]. Though computational simulations like density functional theory (DFT) present accurate approximation of molecular properties, thereby facilitating the engineering of molecules with desired properties such as their solvation energies and electrochemical stability [23, 24, 27], the approach targets the discovery of electrolyte constituents based on their individual properties and neglect to consider their collective impact on the battery's performance as a formulation. Subsequent development of optimum formulation requires brute-force experimentation in shortlisted yet still large chemical space, otherwise, risk getting stuck in local maxima as is the case in most experimental formulation optimization [28]. To address these limitations, we previously published a formulation graph convolution network (F-GCN) model to map the formulation constituent's chemical and structural underpinnings to the overall performance, based on their respective compositions [29]. The model merged constituent screening and optimization into a consolidated effort that can reduce the necessity for heavy computations by relying on limited available battery datasets. By demonstrating good predictive capability for battery electrolytes, the model effectively narrowed favorable electrolyte designs for lithium iodine (Li-$I_2$) battery [29, 30]. Despite the competent representation of complex formulations, the model's scope in driving electrolyte discovery was limited. Since electrolyte is not a stand-alone formulation, but rather, part of a battery device where performance is collectively influenced by cell-level variables like electrode compositions, separator, and current collector. Therefore, ML-driven electrolyte discovery approaches must be extended to incorporate electrolyte optimization based on added cell variables (summarized in Figure 1(a)).

In this work, we demonstrate a data-driven approach summarized in Figure 1(b), that uses an *extended*-FGCN model to enhance the electrolyte performance of an iodine chemistry-based novel interhalogen (Li-ICl) battery. The model maps battery performance to the complex electrolyte formulations while considering additional cell variables such as cathode loading, and separators used. The predictive accuracy of the *extended*-FGCN model is evaluated across a range of cathode loadings, and the results are compared with multiple

ML models. Having the best predictive capability, the trained *extended*-FGCN model (trained with $10^2$ data points) is used to screen high-performing electrolyte formulations for the targeted cathode loadings from a large design space ($10^4 - 10^6$). The screened electrolytes are experimentally validated to demonstrate higher battery capacities, thereby establishing the reliability of the model. Furthermore, an interpretable framework is applied to model predictions for extracting electrolyte design principles custom to the cathode loadings.

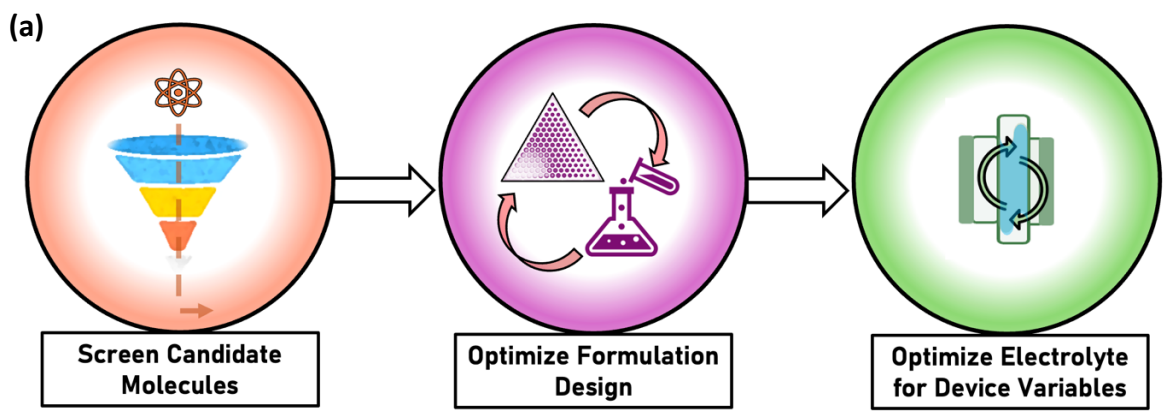

(a) Screen Candidate Molecules → Optimize Formulation Design → Optimize Electrolyte for Device Variables

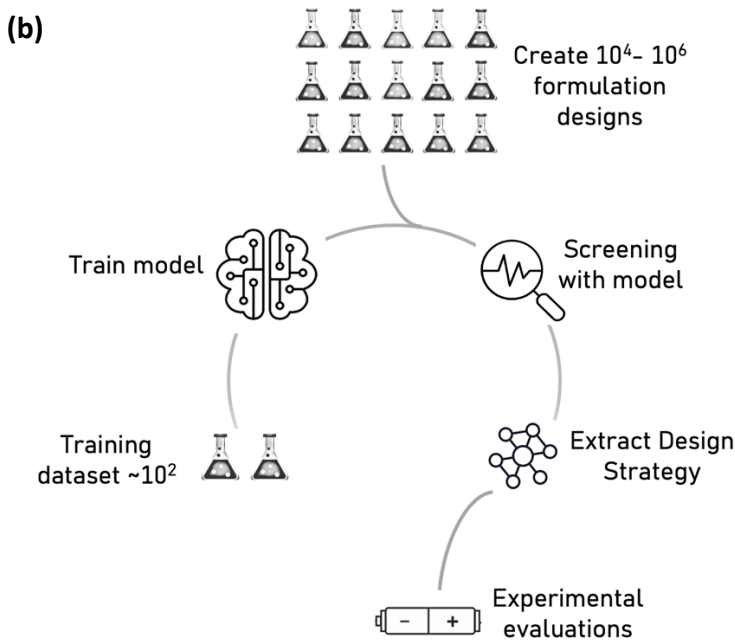

(b) Create $10^4 - 10^6$ formulation designs → Screening with model → Extract Design Strategy → Experimental evaluations → Training dataset ~$10^2$ → Train model

**Figure 1. Electrolyte discovery research workflow. (a)** Electrolyte discovery workflow describing multi-phased research that includes screening candidate molecules, finding optimum formulation design and optimizing electrolytes with respect to device-level variables such as active material loadings and additional cell configuration. **(b)** Schematic summarizing the data-driven approach to extract electrolyte design custom to the cathode loadings.

## 2. Results and Discussion

This section describe the customization of Li-I$_2$ conversion battery to multi-electron redox mediated Li-ICl chemistry by incorporating LiCl salt in the electrolyte design, followed by the experimental and data-driven electrolyte optimization to achieve the best possible Li-ICl battery performance at a desired cathode loading.

### 2.1 Li-ICl Battery Chemistry and Electrolyte Component Selection

Giammona et al.[13] previously published the oxygen-assisted lithium-iodide (OALI) battery (OALI battery), based on LiI conversion chemistry (2Li + I$_2$ ↔ 2LiI) , that demonstrated extremely fast charging capabilities. With its excellent rate capabilities that are desirable for electric vehicle (EV) applications, OALI battery presents an ideal and practical candidate system to deploy interhalogen-driven multi-electron redox reactions [21] for enhancing the overall battery capacity. To achieve this, LiCl salt is introduced into the electrolyte system (Figure 2(a)). The presence of chloride ions (Cl$^-$) instantiates redox chemistry with interhalogen activation of I$^-$/I$^+$ that leads to the formation of an interhalogen charge product (I-Cl) in the OALI battery as per Equations (1-2), where the first electron (e$^-$) transfer happens in Equation (1) and second e⁻ transfers occurs in Equation (2) [21]. This process enhances the theoretical capacity of the battery from 211 mAh/g to 422 mAh/g.

$$I^- - e^- \leftrightarrow I^0 \quad \text{Equation (1)}$$
$$I^0 - e^- + Cl^{-1} \leftrightarrow I^+(Cl^-) \quad \text{Equation (2)}$$

Incorporating LiCl salt to facilitate multi-electron redox chemistry in a OALI battery presents a critical challenge of sparing solubility of LiCl in ether-based solvents which have thus far been demonstrated to be the best solvent system for Li-metal batteries including OALI [13]. Given that at least 1:1 molar ratio of iodide to chloride is required for the formation of I-Cl and the solubility of LiCl is limited to 1-2 molar % in ether-based solvent system (1:1 v/v DOL/DME), excessive electrolyte (> 200 µL/cm² electrode) is needed to achieve the full conversion of iodine to iodine monochloride (I-Cl). With this excess of electrolyte, any gains in energy density from the additional redox chemistry are thus lost to the added electrolyte weight. Therefore, electrolytes with increased chloride solubility should be designed so that high chloride and lean electrolyte conditions can be achieved, simultaneously.

The base electrolyte solvents (1,3-dioxolane (DOL) and tetraglyme (G4)) for the interhalogen variant of OALI battery (thereby referred to as Li-ICl battery) are drawn from our previous publication where solvents were screened for the parent OALI battery [30]. To facilitate LiCl solubility, we shortlist several co-solvent candidates based on their donor number (summarized in Supplemental Table S1). Next, the shortlisted solvent candidates are experimentally investigated for rate capability and cycle life in Li-ICl battery (refer to Supplemental Figure S1). Results indicate that incorporating ethylene carbonate (EC) as a co-solvent in a DOL and G4 (1:1:1 v/v/v) solvent system demonstrates the best capacity and rate capability in Li-ICl battery. Meanwhile, co-solvent 1,3-Dimethyl-2-imidazolidinone (DMI) contributes to the best cycle life. Identifying the complementary significance of both co-solvents to the battery performance, electrolyte design with 4 solvents and 4 salt system is settled upon. The rationale behind the inclusion of each constituent in the electrolyte design is summarized in Supplemental Table S2, and can be the direction of further investigation for these relatively new interhalogen battery chemistries. In present study, we remain focused on the challenging task of finding the optimum electrolyte within the constrained chemical space of 8 molecules with over $10^8$ possible compositional designs.

The specific capacity of Li-ICl battery with varying compositions of 8 electrolyte constituents are measured at a current density of 1 mA/cm$^2$, for the active cathode loadings ranging from 30 wt% to 60 wt% (measured by *LiI wt%* in porous carbon). While anode and coin cell assembly are kept consistent in the experimental data generation, separators used were changed based on varying electrolyte composition to address the issue of wettability. The two separators used are Celgard 2325, which is a trilayer membrane of polypropylene (PP) and polyethylene (PE), and the SiO$_2$-based inorganic QMA. The distribution of battery's specific capacities from 93 experiments is shown in Figure 2(b) across the range of cathode loading. A significant drop in the battery performance with increasing cathode loading can be seen. The highest specific capacity noted is 320 mAh/g at ~32 *LiI wt%* cathode loading. Meanwhile, the Li-ICl battery's capacities upwards of 40 *LiI wt%* are significantly below the expectations. Figure 2(c) compares this performance metric between the parent OALI battery and its interhalogen variant Li-ICl battery at 40 - 46 *LiI wt%* cathode loading. The data on OALI battery (sourced from our previous published studies [29]) is compared with the current data on Li-ICl battery to show a mere 40% increment in the specific capacity (150 mAh/g to 210 mAh/g at ~45 *LiI wt%* ) upon the incorporation of multi-electron redox chemistry. The inverse correlation noted between the battery performance and high cathode loadings is well aligned with the previous studies where it has been reported that increasing the areal capacity of the battery by incrementing active cathode material loading over a critical point can result in a significant drop in the battery's performance across most metrics, primarily due to increased internal resistance, shuttling, and parasitic side reactions [31]. Indeed, several prior studies highlight the challenges associated with increasing active cathode mass loading in $I_2$-based batteries over 32 wt% in carbon substrates [21]. The optimum cathode loading in any battery system is based on several processing and assembly conditions such as calendaring, carbon-to-binder ratio, electrolyte volume, and electrolyte constituents, which require multiple focused optimization efforts [32-34]. The 93 accumulated experimental data points on the Li-ICl battery's performance suggest cathode loadings ranging from 40-45 *LiI wt%* to be the cliff point for the battery performance, beyond which detrimental factors like dissolution of active material in electrolyte and parasitic side reactions might be challenging to control. Therefore, we next adopt a data-driven approach to customize electrolyte formulation in the Li-ICl battery for best performance in this target cathode loading.

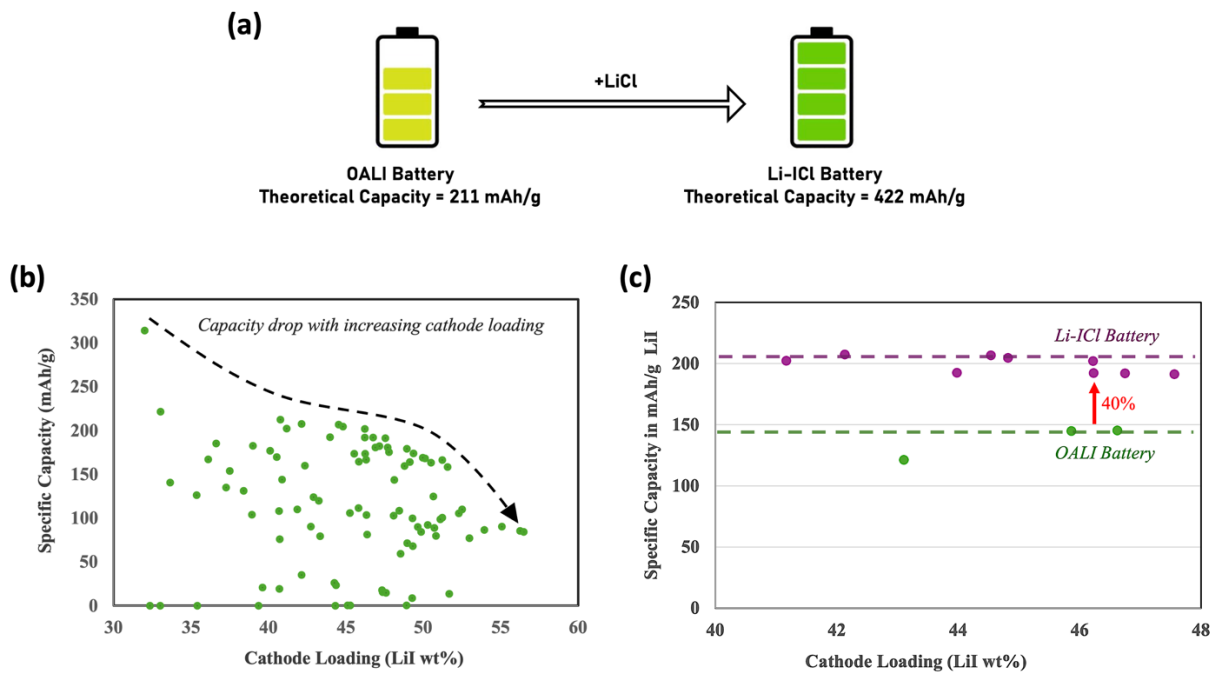

**Figure 2. Performance of Li-ICl battery. (a)** Customization of oxygen-assisted lithium-iodide (OALI) battery to multi-electron redox reaction mediated Li-ICl battery with incorporation of chloride. **(b)** Distribution of Li-ICl battery's specific capacities across a range of cathode loadings from 93 experiments, with varying electrolyte compositions and separators. **(c)** Increment in practical specific capacity of OALI battery [13, 29] with the inclusion of ICl-based multi-electron redox reaction in the optimal cathode loading range.

## 2.2 Battery Capacity Prediction

We train a deep learning model based on F-GCN to map complex electrolyte design to the battery performance based on added cell-level parameters such as cathode loading (*LiI wt%*) and separator (*sep*) type. We previously applied the F-GCN model to map electrolyte constituents to the battery's performance metrics (columbic efficiency and specific capacity), based on their respective molar percentages (*mol%*) in the system [29]. The model tackled the problem of variable chemical constituents and compositions in formulation by scaling each constituent's structural representation by its concentration in *mol%* before additively combining them all into a formulation descriptor. Herein, the F-GCN model is extended (*extended*-FGCN) as shown in Figure 3 to incorporate added cell variables (*LiI wt%* and *sep*) into the framework. Molecular graph representations (*GR*) are derived for 8 electrolyte constituents from graph convolution networks (GCNs) pre-trained with quantum chemical properties (HOMO-LUMO energy levels and electric moment) of the solvent and salt molecules [29]. The GRs from pre-trained GCNs encode information about the molecule's atomic structure and quantum chemical characteristics. Obtained GRs are scaled with each constituent's respective *mol%* in the composition to represent the actual fraction of their chemistry in the formulation. The scaled-GR vectors from 8 constituents are concatenated together along with additional vectors denoting *LiI wt%* and *sep* class. This aggregated layer is transferred to the external learning architecture like dense feed-forward neural networks (DNN), as inputs to map to the final battery performance.

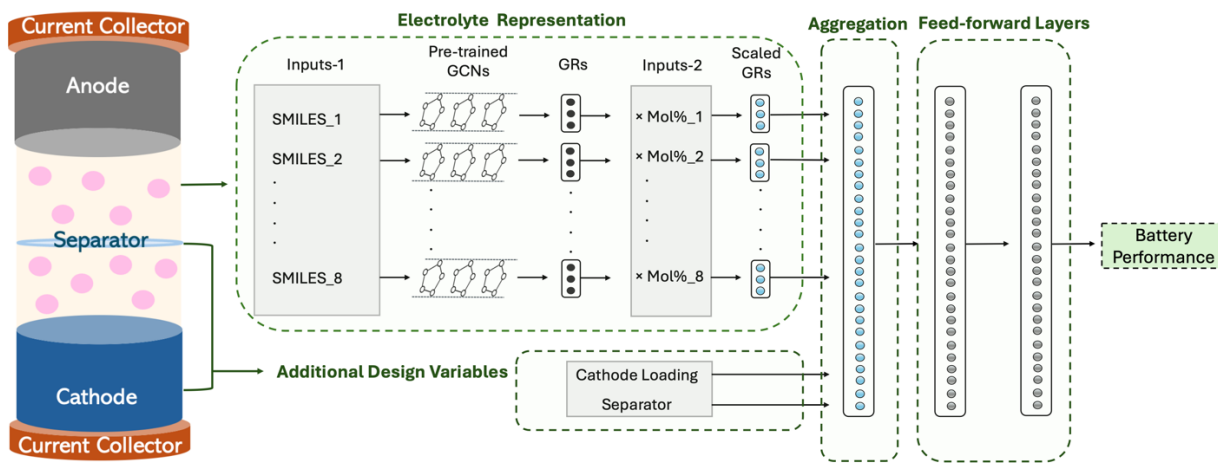

**Figure 3.** *Extended* **Formulation Graph Convolution Network Model.** Schematic representation of extended formulation graph convolution neural networks (*extended*-

FGCN) for mapping electrolyte formulations along with the additional battery cell variables like cathode loading and separator to the battery performance outcome.

The collected experimental dataset with variations in electrolyte composition, cathode loading, and separator in Li-ICl battery is used for model training and evaluation. To assess the accuracies with which the *extended*-FGCN model can associate battery performance to the variations in the electrolyte formulations and cathode loadings, the 93 data points are segregated into training and test datasets for two evaluations based on cathode loadings: thereby referred to as 'random' and 'sorted'. In 'random' evaluation, 20% random data points are set aside as test data points (depicted by orange markers in Figure 4(a)). In 'sorted' evaluation, 20% data points in the high-end cathode loading are set as test data for model evaluation as depicted in Figure 4(b). The 'sorted' dataset is also used to test model hyperparameters summarized in Supplemental Table S3. These evaluations aim to establish the extrapolative capability of the *extended*-FGCN model for cathode loadings. All distributed datasets constitute cells using either of the two separators (Celgard and QMA). The model is trained separately with the two training datasets (random and sorted) and their predictive capabilities are evaluated for their respective test datasets. Parity plots in Figure 4(c-d) summarize the overall results of the two evaluations with the *extended*-FGCN model. Plots indicate the correlation between the specific capacity values predicted by the model and the experimental values. The straight red line in the plots indicates perfect agreement with the experimental capacities and the scatter points denote the predicted capacities for the test electrolyte formulations. The difference of the scatter points from the straight line indicates the errors in the capacity predictions. The errors are evaluated with root mean squared error (RMSE), also specified in Figure 4(c-d). Both models predict specific capacities of cells with RMSE ~25 mAh/g, which is well within the observed experimental cell-to-cell variability in the present dataset (refer to Supplemental Figure S2). Upon further assessment, no clear relationship between the considered variables (electrolyte constituent *mol%*, *LiI wt%* in cathode, and *sep* type) and the data points with relatively large errors is established. Despite few relatively large error points, it is clear from the parity plots that the model is capable of distinguishing high-performing electrolytes.

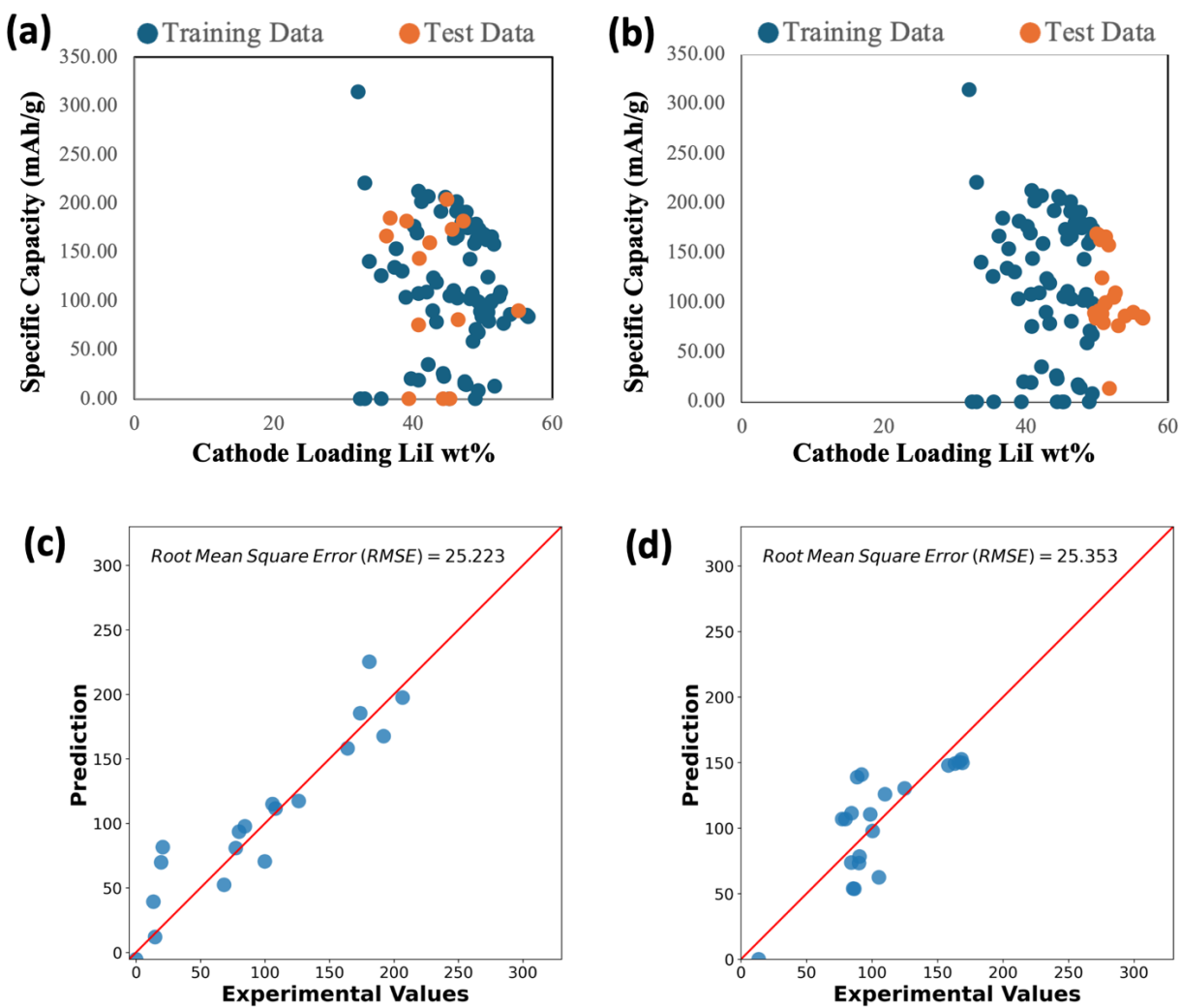

**Figure 4. Model Performance Prediction. (a)** Distribution of specific capacities of electrolytes across cathode loading, and their 'random' segregation into a training dataset

and test data. **(b)** Distribution of specific capacities of electrolytes across cathode loading, and their 'sorted' segregation into a training dataset and test data. Training data are indicated by blue markers and test data are indicated by orange markers. **(c)** Parity plots showing predicted battery capacities (in mAh/g) with respect to the benchmark experimentally derived capacity values mapped on the y=x axis, for 'random' segregated set. **(d)** Parity plots showing predicted battery capacities (in mAh/g) with respect to the benchmark experimentally derived capacity values mapped on the y=x axis, for 'sorted' segregated set.

Since the electrolyte constituent molecules in the present dataset were fixed and only compositions (*mol%*) along with active cathode loading (*LiI wt%*) and separator class (*sep*) were varied, conventional ML models can also be used to map these battery design variables to the performance. The predictive performance of *extended*-FGCN model is compared with popular ML models like Support Vector Regressor (SVR), and Random Forest Regressor (RFR) in Table 1. The results correspond to the best model hyperparameter sets and are presented as mean absolute errors (MAE) for the 'random' train-test (85%-15%) dataset assortment. *Extended*-FGCN demonstrates the lowest MAE (18.79 mAh/g) for the current battery material design dataset, followed by RFR (21.10 mAh/g). The superior performance of the *extended*-FGCN model over the conventional ML models can be attributed to the inefficiency of later models in establishing a true relationship between material compositions to the outcome in the absence of material knowledge. Considering *extended*-F-GCN model retains underlying chemical information of the constituent materials through pre-trained GCNs that featurize molecules on the fly, this model can be an efficient and generalizable acquisition function for Bayesian optimization (BO) methods [26, 35, 36].

**Table 1** Mean Absolute Error (MAE) obtained for Li-ICl battery performance data from ML model based on electrolyte compositions (*mol%*), cathode loading (*LiI wt%*) and separator class

| S.no. | Model | MAE (mAh/g) |
|---|---|---|
| 1. | Support Vector Regression | 30.00 |
| 2. | Random Forest Regression | 21.10 |
| **3.** | *Extended-* FGCN | 18.79 |

## 2.3 Electrolyte Formulation Screening and Validation

Due to the excellent interpolation and extrapolation capability of the *extended*-FGCN model across cathode loadings in section 2.2, we adopt this model to screen formulation designs that demonstrate the best performance for the target cathode loadings. Thus, the model with the best hyperparameter set is trained with all 93 data points. Next, we create a large set (33,740) of dummy battery designs with 2410 electrolyte compositions across 7 cathode loadings that utilize 2 distinct separators. The dummy electrolyte formulations are created randomly to cover all compositional designs equivalently as shown in Supplemental Figure S3, with a small constraint. The formulation designs with a total salt concentration exceeding 50 *mol%* are kept limited to avoid solubility issues. Very high *mol%* of salts will fail to solubilize in the corresponding low *mol%* of the solvents. The trained *extended*-FGCN model is used to predict the specific capacity of 33,740 dummy battery designs, and electrolyte formulations with high predicted capacities for the target cathode loadings (40-*45 LiI wt%*) are screened and validated experimentally. It is important to note that the average predicted capacities for the dummy electrolytes paired with the QMA separator are higher than their use with the Celgard separator (refer to Supplemental Figure S3). Upon further analysis, the correlation between battery performance and separator choice is found to be sensitive to the EC solvent in the system (refer to Supplemental Figure S4). This is potentially due to the increased polarity and viscosity of the EC solvent that prevents it from wetting the surface of a non-polar polymer-based Celgard separator [37, 38]. Based on the suitability of the QMA separator for the present electrolyte design, further electrolyte discovery efforts are limited to battery design using a QMA separator.

From the pool of 2410 formulation designs using QMA, four electrolyte formulations having predicted capacity above the already achieved value of 210 mAh/g are shortlisted (summarized in Table 2). These formulations are validated experimentally by two individuals to account for any uncertainty arising from the manual electrode preparation and cell assembly. Three out of four (3/4) electrolyte formulations successfully demonstrate specific capacities higher than the previous data and are marked in red in Figure 5(a). Due to the error margin in the manual LiI loading process (Supplemental Figure S2), it was challenging to test formulations at a consistent cathode loading. Yet, it was ensured in the series of repetitive experiments that all tested cathode loadings lay within the target 40-45 *LiI wt*% range. The one low-performing formulation (Table 2, *(i)*) was found to have a high concentration of LiCl salt (8 molar %, 0.8 M) that resulted in a cloudy electrolyte (not fully soluble). With the latest validated formulations (indicated in red in Figure 5(a)) demonstrating performance better than the initial dataset (93 data points in green in Figure 5(a)), it can be ascertained that the model retains the ability to differentiate good-performing electrolytes in the target cathode loading range, failing in conditions where the model is oblivious to chemical phenomenon such as collective salt solubility. Thus, the scope of the present model can be further improved with the incorporation of additional solubility knowledge.

**Table 2** Electrolyte formulations in *mol%* from the dummy dataset with the highest predicted capacity in 40-45% cathode loading

| S.no. | LiCl | LiNO$_3$ | LiBOB | LiTFSI | DOL | DMI | EC | G4 |
|---|---|---|---|---|---|---|---|---|
| *(i)* | 8 | 2 | 1 | 3 | 60 | 4 | 14 | 8 |
| *(ii)* | 6 | 2 | 3 | 1 | 64 | 8 | 4 | 12 |
| *(iii)* | 3 | 1 | 1 | 3 | 56 | 8 | 16 | 12 |
| *(iv)* | 4 | 6 | 3 | 1 | 68 | 2 | 10 | 6 |

**2.4 Interpreting Electrolyte Design Custom to Cathode Loading**

We use the trained *extended*-FGCN model to indicate electrolyte formulation design with the high performance at cathode loadings higher than 40 *LiI wt%*. Though large-scale screening with the trained model can assist in identifying suitable formulations, the model alone falls short of providing an intuitive understanding of the electrolyte design for the target battery configuration (anode – cathode materials and compositions). A method that can effectively extract design rules from the model by elaborating the relationship between the variables and target property can add interpretability to the framework. This physical interpretation of correlations can enable further performance improvement. Yet, it remains incredibly challenging to add an interpretability framework to deep learning models such as *extended-*

FGCN where the materials are mapped to the target property through a sequence of descriptor engineering steps: First, the molecular descriptors are derived from a label-dependent pre-training process; next, these descriptors are scaled with their respective compositions, followed by aggregation into a formulation descriptor; and their final mapping to the target performance with a learning architecture. Thus, the latest interpretable and explainable frameworks [39] for deep learning models are not directly relevant to the models customized for complex materials like liquid formulations. Meanwhile, alternative machine learning models that map selected features of the candidate materials to the outcome with comparatively less accuracy can be good candidates for developing generalizable interpretability [40, 41].

To retain the best accuracy and reliability in extracted electrolyte design rules, we adopt Spearman's correlation coefficient (SCC) analysis for interpreting the monotonic association between the electrolyte compositions and battery performance at targeted cathode loadings. Furthermore, to clearly differentiate the material relationships fed to the model during training and relationships learned by the model, we perform separate analyses of training data acquired from experiments (section 2.1) and the dummy dataset used for screening (section 2.3). Due to poor wettability and low cathode loading distribution in the experimental data points utilizing Celgard separator (mostly < 40 *LiI wt*%), only data associated with QMA separator is considered for the analysis. The correlation between battery design variables and outcome capacity in the training data is summarized in Supplemental Table S4. The values closer to 0 indicate no correlation at all, while values closer to +1 and -1 indicate positive and negative correlation, respectively. It is interesting to note that the capacity outcome is barely correlated to the changes in cathode loadings in the range of 40 to 46 *LiI wt%*, with SCC = -0.07. Meanwhile, above 46 LiI wt%, for every 3% change in cathode loadings, there is a very strong negative correlation between battery capacity and cathode loadings (SCC ~ -0.77). These values support our assumption that 40-45 *LiI wt%* cathode loading could be a critical cliff point for Li-ICl battery beyond which other detrimental factors contribute to performance degradation. Since the experimental data points are small in count (53/93 data point with QMA separator), no additional meaningful trend can be established between the variables and the outcome performance, except for a mild positive correlation between capacity and concentrations of LiCl and EC at target cathode loadings.

Next, we perform SCC analysis on the predicted capacities of 2410 dummy electrolyte formulations for a range of cathode loadings using QMA separator. SCC coefficients in Figure 5(b) indicate the correlation between the concentration of each constituent and the predicted capacity in 2410 formulation designs, guiding electrolyte design for high battery performance at different cathode loadings. Correlation analysis is further supported by statistical visualization of the dataset using box plot presented in Supplemental Figure S5 and S6. The electrolyte design principles realized from the analyses of dummy data differed based on cathode loading ranges. For 40 *LiI wt%*, SCC results in Figure 5(b) indicate a positive correlation of battery performance with the concentrations of DOL and a strong negative correlation to the LiBOB salt in the system. Meanwhile, there is a very mild correlation (-0.2< SCC <0.2) with concentrations of other constituents. Specific concentrations of formulation constituents can be visualized in box blots in Supplemental Figure S5. In high-predicted electrolytes, DOL is the major solvent with the highest concentration in the electrolyte (> 50 *mol%*). EC and tetraglyme (G4) are favorable for performance when kept below 15 *mol %*, respectively. Meanwhile, DMI is maintained at a minimum of 1:2 DMI to LiCl ratio. The trend of SCC values for DMI in Figure 5(b) shows an increasing negative correlation with the cathode loadings. Furthermore, reflecting on the solubility issues observed in the screening validation (section 2.3, Table 2 *(i)*), it is advisable to keep LiCl concentration below 7 *mol %* in the electrolyte design, thereby further reducing the need for DMI. Together, both analyses recommend the following electrolyte design:

- Enhance DOL in the system
- Further reduce/ remove LiBOB salt
- Minimize the use of DMI solvent and maintain 1:2 DMI to LiCl ratio

Based on these design recommendations, several electrolyte formulations are tested in coin cell configurations. The specific capacities of the electrolyte formulations validating the extracted design principles are indicated by blue markers in Figure 5(c). Upon experimental evaluation of the latest electrolytes, the removal of LiBOB salt marginally improved the specific capacity of the battery to 250 mAh/g at 1 mA/cm$^2$ for 44 *LiI wt%*. The composition of the electrolyte in *mol%* is as: 4% LiCl, 0% LiBOB, 6% LiNO$_3$, 1% LiTFSI, 68% DOL, 2% DMI, 10% EC, and 9% G4. In addition to all the electrolyte design recommendations mentioned above, the successful formulation also has a very high concentration of SEI stabilizing LiNO$_3$ salt (6 *mol%*). The newly discovered electrolyte formulation also has excellent rate capability and retains 120 mAh/g capacity at 10 mA/cm$^2$ (Supplemental Figure S7). Overall, the data-driven approach leads to an additional ~20% increment in the battery capacity at target cathode loadings (40~45 wt%), after experimental optimization.

This approach also guides electrolyte design rules for lower and higher cathode loadings. For 30 *LiI wt%*, boosting concentrations of EC and DMI while reducing DOL can be beneficial for the battery performance. The correlation trend in Figure 5(b) for lower cathode loading is backed by the singular training data point with a specific capacity higher than 300 mAh/g in Figure 5(a,c). For 60 *LiI wt%,* the SCC values indicate a strong negative correlation to DMI and LiBOB in the system. Additionally, the box plots in Supplemental Figure S6 indicate the favorable concentrations of electrolyte constituents for achieving a battery capacity higher than 150 mAh/g at 60 *LiI wt%.* Though these design principles are extracted from the dataset (dummy data) used for screening high-performing candidates in section 2.3, this additional interpretability provides insights into the understanding developed by the model about the data, and a set of rules for researchers to further evaluate and update the model with feedback.

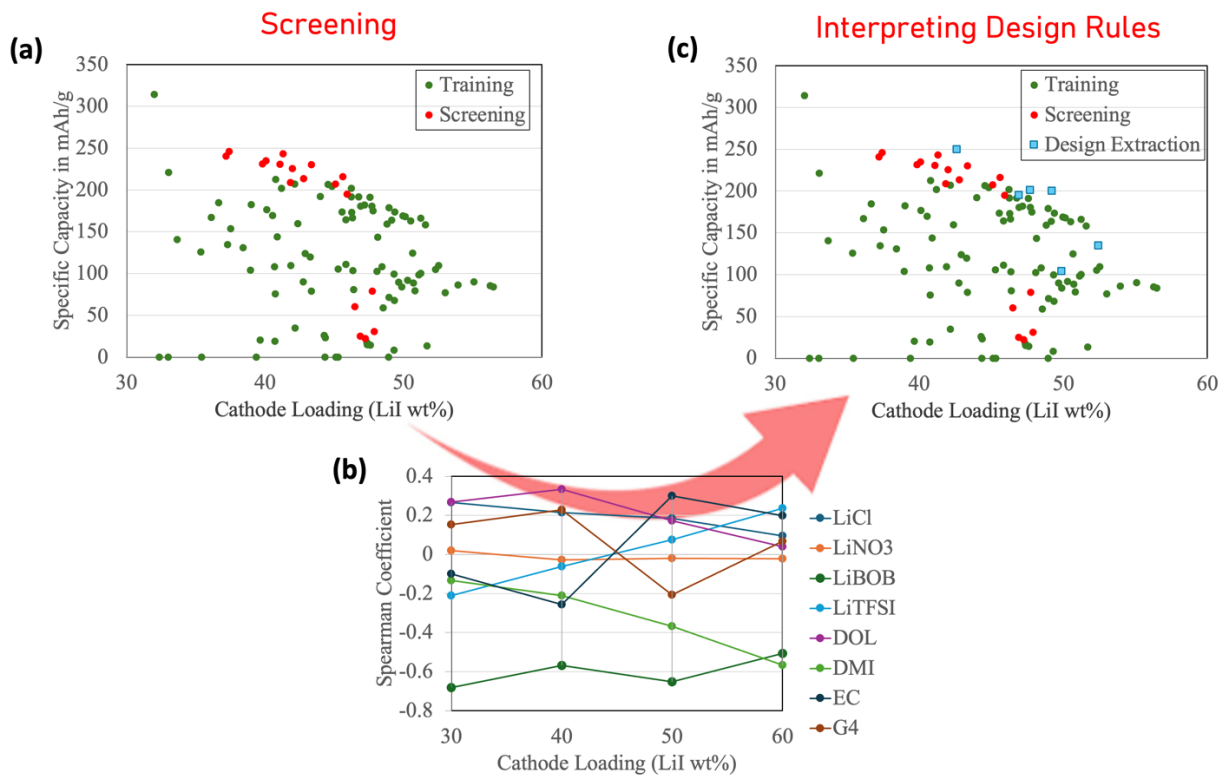

**Figure 5. Experimental validation of screened electrolytes and electrolyte design. (a)** Distribution of experimentally measured specific capacities (in mAh/g) of the electrolytes across cathode loading *LiI wt%* in Li-ICl batteries. The electrolytes in the initial dataset (green marker) are differentiated from new electrolytes discovered by screening (red marker). **(b)** Spearman's coefficient values between electrolyte constituents and battery capacity, were determined from model predictions at different cathode loadings. **(c)** Distribution of experimentally measured specific capacities (in mAh/g) of the electrolytes designed with interpreted rules compared to previous datapoints.

## 3. Summary and Outlook

In summary, we expanded the methodology of data-driven electrolyte discovery to integrate additional battery design variables that evolve with electrolytes synergistically such as active cathode material loading. To accomplish this, the electrolyte of a novel interhalogen Li-ICl battery is experimentally optimized, followed by focused customization of electrolyte for the target cathode loading using a data-driven approach. A graph-based deep learning model (F-GCN) with the ability to map the structure and composition of electrolyte constituents to battery performance is extended to incorporate additional device-level variables (*extended*-FGCN). The model is trained and tested with 93 Li-ICl battery performance data points generated with experiments until the model demonstrates the predictive errors within the bounds of experimental uncertainty. The model is then used to screen electrolyte formulations for high performance at a target cathode loading from the large compositional design space. Upon experimental validation of screened electrolytes, it is established that the model exhibits reliability in screening the high-performing electrolytes. Next, the predicted outcomes from the model for 33,740 battery designs are used to interpret electrolyte design principles custom to different cathode loadings. With the presented data-driven methodology, the specific capacity of the interhalogen-based Li-ICl battery system is enhanced from 210 mAh/g (experimental optimization) to 250 mAh/g (data-driven) at target cathode loading of 45 *LiI wt%,* bringing about ~20% performance improvement. Additionally, interpreting the relationship between the variables and target property provided insights into the understanding developed by the model about the data, and produced a set of electrolyte design rules for different cathode loadings.

Conventionally, with every modification in battery electrodes, electrolytes are revisited for molecular reselection and formulation optimization. However, with the present study, we demonstrate an approach to customize electrolytes for any variations in the electrodes using the available battery datasets that are limited in count (<100) and unconstrained in variables (electrolyte candidates, compositions, electrodes, separator, etc.), which is a common attribute of experimental datasets. The availability of suitable datasets becomes a major constraint for AI-driven material discovery and optimization.

Successful discovery of battery electrolytes requires multi-phased research and optimization at the scale of molecules, mixtures, and devices. While the traditional methods of electrolyte research are laborious, AI-driven research at each scale can stretch the data generation to the point of providing hardly any benefit over the traditional approaches. To leverage the benefits of AI in this domain, models such as *extended*-FGCN that incorporate chemical information of constituent materials, their composition or design, along with the additional device-level variables are desired. The accuracy of the model can be further improved by either incorporating essential domain knowledge such as salt solubilities and interfacial reactions, or utilizing generalizable foundation models for chemistry such as MoLFormer [42].

## 4. Methods

### 4.1 Material Preparation

A composite cathode is used for Li-ICl battery that consists of high surface and high conductivity carbons along with binder and dosed concentration of LiI. For cathode composite preparation, a slurry of microporous super-activated carbon MSC-30SS (Kansai Coke) is prepared with binder styrene-butadiene rubber (SBR, PSBR100, Targray) and carboxymethylcellulose Sodium Salt (CMC, low viscosity, Sigma Aldrich) in water. To enhance conductivity of carbon host, high conductive carbon Ketjen Black (KB, EC600JD, Lion Chemical) is added to the slurry in small quantities as an additive. The slurry contains 10-20 % solids by weight and is spread coated onto a stainless-steel foil (10 um thickness, SS304) that acts as a current collector. The stainless-steel sheets are coated and subsequently dried under ambient conditions. The final composition of the cathode host material on stainless steel is 76.5/8.5/13/2 MSC-30SS/KB/SBR/CMC. This carbon coated stainless steel is then punched into 11 mm ø disks and transferred into an Ar glovebox where un-dosed cathodes are stored on a hotplate at 120 °C until used. Prior to LiI dosing, carbon cathodes have an approximate mass loading of 4-6 mg/cm$^2$ and a thickness of 200 – 300 μm. We used carbon host material prepared in a single batch for the entire study (data collection and validation) to avoid any inconsistencies. However, there are slight variations in the thickness of the cathodes between punched samples that could be accounted for small uncertainties in the experimental measurements. To dose cathodes with active material LiI, carbon disks are weighed individually under an Ar atmosphere followed by dosing of LiI by weight percentage for the target loading (30-60%) from a 1M LiI stock solution made in methanol. The dosed cathodes are dried at 150 °C overnight and then stored in similar conditions until used.

All electrolytes comprise of at most 8 constituents: 4 salts which are Lithium bis(trifluoromethanesulfonyl)imide (LiTFSI, BASF), lithium bisoxolatoborate (LiBOB, Sigma Aldrich), lithium nitrate (LiNO$_3$, Sigma Aldrich) and lithium chloride (LiCl, Sigma Aldrich); and a mixture of 4 solvents containing 1,3-Dioxolane (DOL), 1,3-Dimethyl-2-imidazolidinone (DMI), Ethylene carbonate (EC) and Tetraglyme (G4). All salt and solvent compositions are vastly varied in the electrolytes. All salts are dried at 150 °C for at least a day in an Ar atmosphere prior to use. Simultaneously, all solvents are dried over 3 Å molecular sieves in an Ar atmosphere for at least a day prior to use.

Lithium metal anode as used in the present battery. Li foil (Honjo Metals, 200 μm thickness) is punched to 13 mm ø for 2032 type coin cells. The punched Li anodes are mounted on stainless steel spacers inside an Ar filled glovebox and then double bagged inside resealable pouch bags to be transferred to a clean dry air (CDA) filled dry box for immediate cell assembly. Two different separators with 16 mm diameter are used throughout the study: a Polypropylene/Polyethylene/Polypropylene trilayer membrane (Celgard 2325) and a SiO$_2$ (Grade QMA).

**4.2 Battery Performance Evaluations**

To evaluate electrolyte formulations, 2032 type coin cells were prepared under an atmosphere of clean-dry air (CDA) inside a glove box. Presence of oxygen in the glove box during cell assembly facilitates formation of a solid electrolyte interface (SEI) in OALI battery [13]. Each cell is comprised of a 0.5 mm stainless steel spacer, LiI/carbon cathode coated onto a stainless steel foil, a separator, a lithium metal anode, a second 0.5 mm stainless steel spacer, a wave spring, and 60 μL of electrolyte. All galvanostatic charge and discharge measurements are conducted using EC-Lab® software and a multi-channel potentiostat (BioLogic, VMP3). All potentials are referenced against Li/Li$^+$. Cells are initially cycled 10 times at 0.5 mA/cm$^2$-cathode to form a desirable SEI layer on the Li anode. Formation cycles are followed by rate tests where cells are cycled at different rates for 5 cycles. The average specific capacities at 1 mA/cm$^2$-cathode are taken for the study.

**4.3 Experimental Dataset**

A dataset of a total of 93 electrolytes with variable electrolyte compositions and specific capacities at a current density of 1 mA/cm$^2$ is gathered experimentally for Li-ICl battery coin cells having different *LiI wt%* in cathode and *sep* (either Celgard or QMA). This initial dataset has been generated based on the scientific intuition of a battery researcher. Since LiCl is the major contributor to Li-ICl chemistry, the average concentration of LiCl is highest among the salts in the tested electrolytes. To facilitate LiCl solubilization, 1,3-Dimethyl-2-imidazolidinone (DMI) solvent is consistently present in all evaluated electrolytes. Meanwhile, the remaining three solvent concentrations are explored without any constraints in the dataset ranging from 0% to being a major solvent present in the highest concentration. The distribution of cathode loading from 30% to 60% is also indicated in the box plot (blue) as the measure of *LiI wt%*. The final variable in the battery design is the choice of separator among the polymer-based Celgard and quartz fiber-based QMA. The two separators have very distinct characteristics. The Celgard 2325 separator is a trilayer membrane of polypropylene (PP) and polyethylene (PE) with high-performance expectancy in batteries due to its medium porosity and uniform pore structure [43]. Meanwhile, the QMA separator is an inorganic $SiO_2$-based with a 10-fold increase in thickness. The performance of an electrolyte at a cathode loading (*LiI wt%*) is evaluated by considering the average specific capacity of cells cycled at a current density of 1 $mA/cm^2$. The experimental dataset inherits some uncertainties based on slight unintended variations in the electrode preparations. The standard deviation (cell-to-cell variation) of experimental specific capacities is in the range of 15-30 mAh/g (see Supplemental Figure S3) when the same electrolyte formulation is used in multiple coin cell batteries.

## 4.4 Extended Formulation Graphs

Formulation graph convolution networks (F-GCN) [29], a deep learning model to map chemical constituents to a formulation property based on their respective compositions, is extended to incorporate new cell-level variables in addition to the electrolyte formulation. Each electrolyte constituent is initially represented by SMILES (Simplified Molecular Input Line Entry System) [44] and subsequently converted to their respective molecular conformations using an RDKit package [45]. The structural conformations are transformed into molecular graphs consisting of a node matrix embedding atomic configurations of the molecule and an adjacency matrix describing the atomic bonds. Pre-trained GCN are used to convert molecular graphs to a 1-dimensional vector of size 100 called Graph Representations (*GR*) which encodes structural characteristics of the molecules as the function of corresponding quantum chemical attributes (like HOMO-LUMO, Electric Moment) [29]. The GR of each molecule is scaled with the molar percentage of the respective molecule in the formulation to represent actual fraction of constituent components in formulation. All scaled GRs (size 100 each) for eight electrolyte components are concatenated along with additional cell-level input variables such as *LiI wt%* in cathode and separator class (with Celgard represented by class-*1* and QMA represented by class-*2*). This concatenated battery descriptor is further mapped as input layer to a feed-forward neural network for learning battery performance outcome such as specific capacity. It is realized in our previous study that formulation models targeting battery electrolytes require custom hyper-parameter tuning for every performance outcome (capacity, columbic efficiency). While the base GCN for molecule representation remain consistent, the aggregator applied to form a formulation descriptor and external learning architecture may vary based on the dataset and output label. For the present dataset comprising of fixed 8 constituents and variable compositions, concatenation of formulants GRs along with cell-level variables gave best results when paired with the feed forward neural network containing 3 hidden layers with each consisting of 1000, 500 and 100 nodes, as compared to previous applied aggregators [29]. The performance of the model with different hyperparameters is compared in Supplemental Table S3. This customized version of F-GCN for cell level description is referred to as *extended*-FGCN.

## 4.5. Model Training

A Python version of the *extended*-FGCN model is developed using the *Keras* API [46] with TensorFlow [47]. The loss is calculated by mean squared error (MSE) during the training process. Model is trained using Adam optimizer [48] with a batch size 1 and a step-learning rate (lr). After first 4000 epochs at lr = $10^{-4}$, lr is increased by a fraction ($10^{-3}$ to $10^{-2}$) for every next 3000 epochs until the training is stopped. The complete convergence of the model is avoided to escape model overfitting by early stopping process. The 20% segregated test data is also used for validation during the training process. Model gave lowest RMSE for the test data when the loss MSE is ∼ 20 mAh/g.

## AUTHOR CONTRIBUTIONS

M.G., and Y.H.L. conceptualized the project. V.S. developed the model, and performed coding, training, model predictions, and dummy data analysis. A.T. and K.N. contributed equally to this work. Both authors generated experimental datasets and performed electrolyte validation. M.Z. performed data conversion and data analysis. L.S. contributed to material preparation for the battery.

## CONFLICT OF INTEREST STATEMENT

The authors have no conflicts of interest to declare. All authors have seen and agree with the contents of the manuscript and there is no financial interest to report. We certify that the submission is original work and is not under review at any other publication.

**DATA AND CODE AVAILABILITY**

All datasets and codes used in the study will be uploaded in a public GitHub repository.

**REFERENCES**

[1] G. Pilania, C. Wang, X. Jiang, S. Rajasekaran, R. Ramprasad, Accelerating materials property predictions using machine learning, Scientific reports 3(1) (2013) 2810.
[2] N. Meftahi, M. Klymenko, A.J. Christofferson, U. Bach, D.A. Winkler, S.P. Russo, Machine learning property prediction for organic photovoltaic devices, npj computational materials 6(1) (2020) 166.
[3] J. Westermayr, J. Gilkes, R. Barrett, R.J. Maurer, High-throughput property-driven generative design of functional organic molecules, Nature Computational Science 3(2) (2023) 139-148.
[4] M. Manica, J. Born, J. Cadow, D. Christofidellis, A. Dave, D. Clarke, Y.G.N. Teukam, G. Giannone, S.C. Hoffman, M. Buchan, Accelerating material design with the generative toolkit for scientific discovery, npj Computational Materials 9(1) (2023) 69.
[5] P. Liu, H. Huang, S. Antonov, C. Wen, D. Xue, H. Chen, L. Li, Q. Feng, T. Omori, Y. Su, Machine learning assisted design of γ′-strengthened Co-base superalloys with multi-performance optimization, npj Computational Materials 6(1) (2020) 62.
[6] S. Wang, Y. Huang, W. Hu, L. Zhang, Data-driven optimization and machine learning analysis of compatible molecules for halide perovskite material, npj Computational Materials 10(1) (2024) 114.
[7] G. Turon, J. Hlozek, J.G. Woodland, A. Kumar, K. Chibale, M. Duran-Frigola, First fully-automated AI/ML virtual screening cascade implemented at a drug discovery centre in Africa, Nature Communications 14(1) (2023) 5736.
[8] S. Huang, J.M. Cole, A database of battery materials auto-generated using ChemDataExtractor, Scientific Data 7(1) (2020) 260.
[9] K. Li, J. Wang, Y. Song, Y. Wang, Machine learning-guided discovery of ionic polymer electrolytes for lithium metal batteries, nature communications 14(1) (2023) 2789.
[10] J. Xie, Y. Zhou, M. Faizan, Z. Li, T. Li, Y. Fu, X. Wang, L. Zhang, Designing semiconductor materials and devices in the post-Moore era by tackling computational challenges with data-driven strategies, Nature Computational Science (2024) 1-12.
[11] A. Merchant, S. Batzner, S.S. Schoenholz, M. Aykol, G. Cheon, E.D. Cubuk, Scaling deep learning for materials discovery, Nature 624(7990) (2023) 80-85.

[12] F. Rahmanian, R.M. Lee, D. Linzner, K. Michel, L. Merker, B.B. Berkes, L. Nuss, H.S. Stein, Attention towards chemistry agnostic and explainable battery lifetime prediction, npj Computational Materials 10(1) (2024) 100.
[13] M.J. Giammona, J. Kim, Y. Kim, P. Medina, K. Nguyen, H. Bui, G.O. Jones, A.T. Tek, L. Sundberg, A. Fong, Oxygen Assisted Lithium-Iodine Batteries: Towards Practical Iodine Cathodes and Viable Lithium Metal Protection Strategies, Advanced Materials Interfaces 10(17) (2023) 2300058.
[14] Y. Wang, Q. Sun, Q. Zhao, J. Cao, S. Ye, Rechargeable lithium/iodine battery with superior high-rate capability by using iodine–carbon composite as cathode, Energy & Environmental Science 4(10) (2011) 3947-3950.
[15] Q. Zhao, Y. Lu, Z. Zhu, Z. Tao, J. Chen, Rechargeable lithium-iodine batteries with iodine/nanoporous carbon cathode, Nano letters 15(9) (2015) 5982-5987.
[16] L. Qiao, C. Wang, X.S. Zhao, Encapsulation of iodine in nitrogen-containing porous carbon plate arrays on carbon fiber cloth as a freestanding cathode for lithium-iodine batteries, ACS Applied Energy Materials 4(7) (2021) 7012-7019.
[17] Q. Zhang, Z. Wu, F. Liu, S. Liu, J. Liu, Y. Wang, T. Yan, Encapsulating a high content of iodine into an active graphene substrate as a cathode material for high-rate lithium–iodine batteries, Journal of materials chemistry A 5(29) (2017) 15235-15242.
[18] M. Zhao, B.Q. Li, H.J. Peng, H. Yuan, J.Y. Wei, J.Q. Huang, Lithium–sulfur batteries under lean electrolyte conditions: challenges and opportunities, Angewandte Chemie International Edition 59(31) (2020) 12636-12652.
[19] P. Mu, T. Dong, H. Jiang, M. Jiang, Z. Chen, H. Xu, H. Zhang, G. Cui, Crucial challenges and recent optimization Progress of metal–sulfur battery electrolytes, Energy & Fuels 35(3) (2021) 1966-1988.
[20] E.M. Erickson, E. Markevich, G. Salitra, D. Sharon, D. Hirshberg, E. de la Llave, I. Shterenberg, A. Rosenman, A. Frimer, D. Aurbach, Development of advanced rechargeable batteries: a continuous challenge in the choice of suitable electrolyte solutions, Journal of The Electrochemical Society 162(14) (2015) A2424.
[21] X. Li, Y. Wang, Z. Chen, P. Li, G. Liang, Z. Huang, Q. Yang, A. Chen, H. Cui, B. Dong, Two-Electron Redox Chemistry Enabled High-Performance Iodide-Ion Conversion Battery, Angewandte Chemie International Edition 61(9) (2022) e202113576.
[22] X. Qu, A. Jain, N.N. Rajput, L. Cheng, Y. Zhang, S.P. Ong, M. Brafman, E. Maginn, L.A. Curtiss, K.A. Persson, The Electrolyte Genome project: A big data approach in battery materials discovery, Computational Materials Science 103 (2015) 56-67.
[23] L. Cheng, R.S. Assary, X. Qu, A. Jain, S.P. Ong, N.N. Rajput, K. Persson, L.A. Curtiss, Accelerating electrolyte discovery for energy storage with high-throughput screening, The journal of physical chemistry letters 6(2) (2015) 283-291.
[24] M.D. Halls, K. Tasaki, High-throughput quantum chemistry and virtual screening for lithium ion battery electrolyte additives, Journal of Power Sources 195(5) (2010) 1472-1478.
[25] A. Narayanan Krishnamoorthy, C. Wölke, D. Diddens, M. Maiti, Y. Mabrouk, P. Yan, M. Grünebaum, M. Winter, A. Heuer, I. Cekic-Laskovic, Data-Driven Analysis of High-Throughput Experiments on Liquid Battery Electrolyte Formulations: Unraveling the Impact of Composition on Conductivity, Chemistry-Methods 2(9) (2022) e202200008.

[26] A. Dave, J. Mitchell, S. Burke, H. Lin, J. Whitacre, V. Viswanathan, Autonomous optimization of non-aqueous Li-ion battery electrolytes via robotic experimentation and machine learning coupling, Nature communications 13(1) (2022) 5454.

[27] X. Fan, C. Wang, High-voltage liquid electrolytes for Li batteries: progress and perspectives, Chemical Society Reviews 50(18) (2021) 10486-10566.

[28] A. Benayad, D. Diddens, A. Heuer, A.N. Krishnamoorthy, M. Maiti, F.L. Cras, M. Legallais, F. Rahmanian, Y. Shin, H. Stein, High-throughput experimentation and computational freeway lanes for accelerated battery electrolyte and interface development research, Advanced Energy Materials 12(17) (2022) 2102678.

[29] V. Sharma, M. Giammona, D. Zubarev, A. Tek, K. Nugyuen, L. Sundberg, D. Congiu, Y.-H. La, Formulation graphs for mapping structure-composition of battery electrolytes to device performance, Journal of Chemical Information and Modeling 63(22) (2023) 6998-7010.

[30] B. Elmegreen, H.F. Hamann, B. Wunsch, T. Van Kessel, B. Luan, T. Elengikal, M. Steiner, R.N.B. Ferreira, R.L. Ohta, F.L. Oliveira, MDLab: AI frameworks for carbon capture and battery materials, Frontiers in Environmental Science (2023).

[31] Y. Kim, M. Kim, T. Lee, E. Kim, M. An, J. Park, J. Cho, Y. Son, Investigation of mass loading of cathode materials for high energy lithium-ion batteries, Electrochemistry Communications 147 (2023) 107437.

[32] G. Lenze, H. Bockholt, C. Schilcher, L. Froböse, D. Jansen, U. Krewer, A. Kwade, Impacts of variations in manufacturing parameters on performance of lithium-ion-batteries, Journal of The Electrochemical Society 165(2) (2018) A314.

[33] D. Lv, J. Zheng, Q. Li, X. Xie, S. Ferrara, Z. Nie, L.B. Mehdi, N.D. Browning, J.G. Zhang, G.L. Graff, High energy density lithium–sulfur batteries: challenges of thick sulfur cathodes, Advanced Energy Materials 5(16) (2015) 1402290.

[34] M.J. Giammona, J. Kim, Y. Kim, P. Medina, K. Nguyen, H. Bui, G.O. Jones, A.T. Tek, L. Sundberg, A. Fong, Oxygen Assisted Lithium-Iodine Batteries: Towards Practical Iodine Cathodes and Viable Lithium Metal Protection Strategies, Advanced Materials Interfaces (2023) 2300058.

[35] S.G. Baird, J.R. Hall, T.D. Sparks, Compactness matters: Improving Bayesian optimization efficiency of materials formulations through invariant search spaces, Computational Materials Science 224 (2023) 112134.

[36] A.Z. Hezave, M. Lashkarbolooki, S. Raeissi, Using artificial neural network to predict the ternary electrical conductivity of ionic liquid systems, Fluid Phase Equilibria 314 (2012) 128-133.

[37] J.D. Evans, Y. Sun, P.S. Grant, Sequential Deposition of Integrated Cathode–Inorganic Separator–Anode Multilayers for High Performance Li-Ion Batteries, ACS Applied Materials & Interfaces 14(30) (2022) 34538-34551.

[38] M. He, X. Zhang, K. Jiang, J. Wang, Y. Wang, Pure inorganic separator for lithium ion batteries, ACS applied materials & interfaces 7(1) (2015) 738-742.

[39] A.M. Salih, Z. Raisi-Estabragh, I.B. Galazzo, P. Radeva, S.E. Petersen, K. Lekadir, G. Menegaz, A Perspective on Explainable Artificial Intelligence Methods: SHAP and LIME, Advanced Intelligent Systems (2024) 2400304.

[40] H. Choubisa, P. Todorović, J.M. Pina, D.H. Parmar, Z. Li, O. Voznyy, I. Tamblyn, E.H. Sargent, Interpretable discovery of semiconductors with machine learning, npj Computational Materials 9(1) (2023) 117.

[41] J. Dean, M. Scheffler, T.A. Purcell, S.V. Barabash, R. Bhowmik, T. Bazhirov, Interpretable machine learning for materials design, Journal of Materials Research 38(20) (2023) 4477-4496.
[42] J. Ross, B. Belgodere, V. Chenthamarakshan, I. Padhi, Y. Mroueh, P. Das, Large-scale chemical language representations capture molecular structure and properties, Nature Machine Intelligence 4(12) (2022) 1256-1264.
[43] X. Huang, Separator technologies for lithium-ion batteries, Journal of Solid State Electrochemistry 15(4) (2011) 649-662.
[44] D. Weininger, SMILES, a chemical language and information system. 1. Introduction to methodology and encoding rules, Journal of chemical information and computer sciences 28(1) (1988) 31-36.
[45] G. Landrum, RDKit: A software suite for cheminformatics, computational chemistry, and predictive modeling, Greg Landrum (2013).
[46] F. Chollet, Keras: The python deep learning library, Astrophysics source code library (2018) ascl: 1806.022.
[47] M. Abadi, P. Barham, J. Chen, Z. Chen, A. Davis, J. Dean, M. Devin, S. Ghemawat, G. Irving, M. Isard, {TensorFlow}: a system for {Large-Scale} machine learning, 12th USENIX symposium on operating systems design and implementation (OSDI 16), 2016, pp. 265-283.
[48] Z. Zhang, Improved adam optimizer for deep neural networks, 2018 IEEE/ACM 26th International Symposium on Quality of Service (IWQoS), IEEE, 2018, pp. 1-2.